\documentclass[11pt,a4paper]{article}
\usepackage[hyperref]{acl2020}
\usepackage{times}
\usepackage{latexsym}

\usepackage{microtype}

\usepackage{graphicx}
\usepackage{amsmath}
\usepackage{subcaption} 
\usepackage{textcomp} 
\usepackage{multicol} 
\usepackage{multirow}
\usepackage[inline]{enumitem} 

\aclfinalcopy 

\title{Distilling neural networks into skipgram-level decision lists}

\author{Madhumita Sushil\textsuperscript{1} \and Simon \v{S}uster\textsuperscript{2,}\thanks{\hspace{0.3em} Research conducted while at CLiPS.} \and Walter Daelemans\textsuperscript{1} \\
  \textsuperscript{1} Computational Linguistics and Psycholinguistics Research Center (CLiPS), \\
  University of Antwerp, Belgium \\
  \texttt{firstname.lastname@uantwerpen.be} \\ \\
  \textsuperscript{2} School of Computing and Information Systems, University of Melbourne \\ 
    \texttt{simon.suster@unimelb.edu.au}}

\date{}

\begin{document}
\maketitle
\begin{abstract}
Several previous studies on explanation for recurrent neural networks focus on approaches that find the most important input segments for a network as its explanations. In that case, the manner in which these input segments combine with each other to form an explanatory pattern remains unknown. To overcome this, some previous work tries to find patterns (called rules) in the data that explain neural outputs. However, their explanations are often insensitive to model parameters, which limits the scalability of text explanations. To overcome these limitations, we propose a pipeline to explain RNNs by means of decision lists (also called rules) over skipgrams. For evaluation of explanations, we create a synthetic sepsis-identification dataset, as well as apply our technique on additional clinical and sentiment analysis datasets. We find that our technique persistently achieves high explanation fidelity and qualitatively interpretable rules.

\end{abstract}

\section{Introduction}

Understanding and explaining decisions of complex models such as neural networks has recently gained a lot of attention for engendering trust in these models, improving them, and understanding them better~\citep{MONTAVON20181,DBLP:journals/corr/abs-1904-04063,belinkov-glass-2019-analysis}. Several previous studies developing interpretability techniques provide a set of input features or segments that are the most salient for the model output. Approaches such as input perturbation and gradient computation are popular for this purpose~\citep{ancona2018towards,arras-etal-2019-evaluating}. A drawback of these approaches is the lack of information about interaction between different features. While heatmaps~\citep{DBLP:journals/corr/LiMJ16a,li-etal-2016-visualizing,Arras2017WhatIR} and partial dependence plots~\citep{lundberg2017unified} are popularly used, they only provide a qualitative view which quickly gets complex as the number of features increases. To overcome this limitation, rule induction for model interpretability has become popular, which accounts for interactions between multiple features and output classes~\citep{lakkaraju2017interpretable,DBLP:journals/corr/PuriGAVK17,ming2018rulematrix,anchors:aaai18,W18-5411,Evans:2019:WIB:3321707.3321726,Pastor:2019:EBB:3297280.3297328}. Most of these work treat the explained models as black boxes, and fit a separate interpretable model on the original input to find rules that mimic the output of the explained model. However, because the interpretable model does not have information about the parameters of the complex model, global explanation is expensive, and the explaining and explained models could fit different curves to arrive to the same output.~\citet{W18-5411} incorporate model gradients in the explanation process to overcome these challenges, but their technique cannot be used with current state-of-the-art models that use word embeddings due to their reliance on interpretable model input in the form of bag-of-words.~\citet{DBLP:conf/iclr/MurdochS17} explain long short term memory networks (LSTMs)~\citep{hochreiter1997long} by means of ngram rules, but their rules are limited to presence of single ngrams and do not capture interaction between ngrams in text. To learn explanation rules for RNNs while overcoming the limitations of the previous approaches, we have the following contributions in the paper: 
\begin{enumerate}
    \item We induce explanation rules over important skipgrams in text, while ensuring that these rules generalize to unseen data. To this end, we quantify skipgram importance in LSTMs by first pooling gradients across embedding dimensions to compute word importance, and thereby aggregating them into skipgram importance. Skipgrams incorporate word order in explanations and increase interpretability. 
    \item To overcome existing limitations with automated explanation evaluation~\citep{lertvittayakumjorn2019human,P18-1032}, we provide a synthetic clinical text classification dataset for evaluating interpretability techniques. We construct this dataset according to existing medical knowledge and clinical corpus. We validate our explanation pipeline on this synthetic dataset by recovering the labeling rules of the dataset. We then apply our pipeline to two clinical datasets for sepsis classification, and one dataset for sentiment analysis. We confirm that the explanation results obtained on synthetic data are scalable to real corpora.
\end{enumerate}    

\section{Explanation pipeline}

\begin{figure}
\centering
\includegraphics[width=0.97\linewidth]{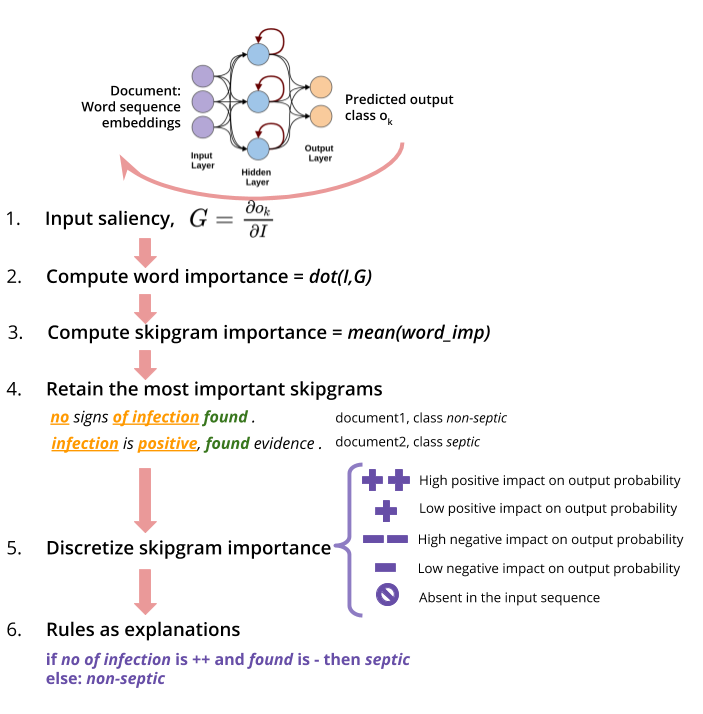}
\caption{The complete \textsc{unravel} pipeline for gradient-informed rule induction in recurrent neural networks. The underlined terms in point 4 refer to different important skipgrams in the text.}
\label{fig:pipeline}
\end{figure}

We propose a method to find decision lists as explanation rules for RNNs with word embedding input. We quantify word importance in an RNN by comparing multiple pooling operations (qualitatively and quantitatively). After establishing a desired pooling technique, we move to finding importance of skipgrams, which provides larger context around words in explanations. We then find decision lists that associate the relative importance of multiple skipgrams in the RNN to an output class. The most similar work to ours is by \citet{W18-5411} who find if-then-else rules for feedforward neural networks. However, their approach relies on using interpretable inputs independent of word order and is not scalable to the current state-of-the-art approaches that use word embeddings instead. Moreover, explanation of binary classifiers is not supported by their pipeline, and their explanation rules are not generalized to unseen examples. Furthermore, their explanation rules are hierarchical, and hence cannot be understood independently without parsing the entire rule hierarchy. In our work, we address all these limitations and extend the explanations to binary cases, unseen data, and to sequential neural networks with word embedding input. Additionally, our explanation rules can be understood as an independent decision path. We present the complete pipeline for our approach, which we name \textsc{unravel}, in Figure~\ref{fig:pipeline}. 
Code for the paper is available on \url{http://github.com/}.

\subsection{Word importance computation}
\label{sec:pooling}

Saliency (importance) scores of input features are often computed as gradients of the predicted output node w.r.t.\ all the input nodes for all the instances~\citep{simonyan2013deep,Adebayo:2018:SCS:3327546.3327621}. In neural architectures that have an embedding layer, interpretable input features are replaced by corresponding low-dimensional embeddings. Due to this, we obtain different saliency scores for different embedding dimensions of a word in a document. Because embedding dimensions are not interpretable, it is difficult to understand what these multiple saliency scores mean. To instead obtain a single score for a word by combining saliency values of all the dimensions, we consider the following commonly used pooling techniques:

\begin{itemize}
    \item 
    \textbf{L2 norm} of the gradient scores~\citep{Bansal:2016:AGM:2959100.2959180,Hechtlinger2016InterpretationOP,P18-1032}.
    $$ saliency_{L2} = \Sigma_{dim}grad^2  $$
    
    \item 
    \textbf{Sum} of gradients across all the dimensions. 
    $$ saliency_{sum} = \Sigma_{dim}grad $$
    
    \item 
    \textbf{Dot product} between the embeddings and the gradient scores~\citep{Denil2014c,MONTAVON20181,arras-etal-2019-evaluating}. This additionally accounts for the embedding value itself.
    $$ saliency_{dot} = \Sigma_{dim}(emb \odot grad) $$
    
\end{itemize}

In Section~\ref{sec:res_pooling}, we analyze the importance scores obtained with these techniques qualitatively and quantitatively to identify the preferred one.

\subsection{Skipgrams to incorporate context}
\label{sec:sg}

One of the contributions of this work is to find explanation rules for sequential models such as RNNs. Conjunctive clauses of if-then-else rules are order independent although this order is critical for RNNs. To account for word order in input documents, some previous approaches find the most important ngrams instead of the top words only~\citep{DBLP:conf/iclr/MurdochS17,jacovi-etal-2018-understanding}. To incorporate word order also in explanation rules, we compute the importance of subsequences in the documents before combining different subsequences into conjunctive rules. We define importance of a subsequence as the mean saliency of all the tokens in that subsequence. We represent subsequences as skipgrams with length in the range [1,4] and with maximum two skip tokens\footnote{Length and skip values in skipgrams were manually decided to include sufficient context while limiting complexity.}. After computing the scores, we retain the 50 most important skipgrams for every document (based on absolute importance scores). The number of unique skipgrams obtained in this manner is very high. To limit the complexity of explanations, we retain 5k skipgrams with the highest total absolute importance score across the entire training set and learn explanation rules over these. To this end, we create a bag-of-skipgram-importance representation of the documents, where the vocabulary corresponds to the 5k most important skipgrams across the training set. For ease of understanding, we discretize the importance scores of the skipgrams to represent five different levels of importance: \{$--$, $-$, $0$, $+$, $++$\}. Here $--$ and $++$ represent a high negative and positive importance, respectively, for the predicted output class, $0$ means that the skipgram is absent in the document, and $-$ and $+$ indicate low negative and positive importance scores, respectively. This skipgram set, along with the output predictions of a model, is then input to a rule induction module to obtain decision lists as explanations. 

\subsection{Learning transferable explanations}
\label{sec:rule_induction}

In the prediction phase, a model merely applies the knowledge it has learned from the training data. Hence, an explanation technique should not require prior knowledge of the test set to find global explanations of a model. We hypothesize that explanation rules should be consistently accurate between the training data and the predictions on unseen data. In accordance to this hypothesis, instead of learning explanations directly from validation or test instances, which is common in interpretability research \citep{anchors:aaai18,W18-5411}, we modify the explanation procedure to learn accurate, transferable explanations only from the training set. We first feed the training data to our neural network and record the corresponding output predictions. These output predictions, combined with the corresponding set of top discretized skipgrams, are used to fit the rule inducer. The hyperparameters of the rule inducer are optimized to best explain the validation set outputs. Finally, we report a score that quantifies how well the learned rules transfer to the test predictions. This training scheme ensures that the explanations are generalizable to unseen data, instead of overfitting the test set.

We obtain decision lists using PART~\citep{Frank1998}, which finds simplified paths of partial C4.5 decision trees. These decision lists can be comprehended independent of the order, and support both binary and multi-class cases. 

\section{Datasets and Models}

\subsection{Synthetic dataset}
A big challenge for interpretability research is the evaluation of the results~\citep{lertvittayakumjorn2019human}. Human evaluation is not ideal because a model can learn correct classification patterns that are counter-intuitive for humans~\cite{P18-1032}. In complex domains like healthcare, such an evaluation is additionally infeasible. To overcome existing limitations with automated evaluation of explanations, we create a synthetic binary clinical document classification dataset. We base the dataset construction on the sepsis screening guidelines\footnote{\url{https://bit.ly/3575e3d}}. 
This is a critical task for preventing deaths in ICUs~\citep{DBLP:conf/icml/FutomaHH17} and new insights about the problem are important in the medical domain. The synthetic dataset includes a subset of sentences from the freely available clinical corpus MIMIC-III~\citep{johnson2016mimic}. Dataset construction process is described here:

\begin{itemize}
    \item From the MIMIC-III corpus, we sample 3--15 words long sentences that mention the keywords discussed in the screening guidelines, grouped into the following sets:
    \begin{enumerate}
        \item \label{set:infection}
        $I$: Contains sentences that mention these infection-related keywords: \{pneumonia \textit{and}\footnote{Sentences mentioning both the keywords are sampled.} empyema, meningitis, endocarditis, infection\}.
        \item \label{set:infl_response}
        $Infl$:  Contains sentences that mention these inflammation-related keywords: \{hypothermia \textit{or}\footnote{Sentences mentioning either of the keywords are sampled.} hyperthermia, leukocytosis \textit{or} leukopenia, altered mental status, tachycardia, tachypnea, hyperglycemia\}.
        \item \label{set:negative}
        $Others$: Sentences that do not mention any of the previously stated keywords: $Sentence \notin\{I \cup Infl\}$.
    \end{enumerate}
    
    \item We populate 50k documents with 17 sentences each by randomly sampling one sentence from set $I$, one sentence for each comma-separated term in set $Infl$, and 10 sentences from set $Others$. We additionally populate 20k documents with 17 sentences, all from set $Others$.
    
    \item We then run the \textsc{clamp} clinical NLP pipeline~\citep{doi:110.1093/jamia/ocx132} to identify if these keywords are negated in the documents. 
    
    \item Next, we assign class labels to the documents using the following rule:
    
    \begin{quote} 
    \textbf{if} \textit{the infection term sampled from set $I$ is not negated} \textbf{and} \textit{at least 2 responses sampled from set $Infl$ are not negated} \newline
    \hspace{20em} $\implies$ Class label is \textit{septic}, \newline
    Class label is \textit{non-septic} otherwise. 
    \end{quote}
\end{itemize}

\noindent 49\% of the documents are thus labeled as \textit{septic}. 

Sampling sentences from the MIMIC-III corpus introduces language diversity through a large vocabulary and varied sentence structures. Use of an imperfect tool to identify negation for document labeling also adds noise to the dataset. These properties are desirable because they allow for controlled explanation evaluation while also simulating real world corpora and tasks, unlike several synthetic datasets used for explanation evaluation~\citep{arras-etal-2019-evaluating,DBLP:conf/acl/ChrupalaA19}.

\subsubsection{Gold important terms}
\label{sec:gold}

For every document, the set of words that are used to assign it a class label includes all the keyword terms about infection from set $I$ that are mentioned in that document, keyword terms about inflammatory response from set $Infl$, and their corresponding negation markers as identified by the \textsc{clamp} pipeline. We mark these sets of terms, one set per document, as the gold set of important terms for this task. For example in the document:

\begin{quote} 
\underline{No} \underline{signs} \underline{of} \underline{infection} were found. \underline{Altered} \underline{mental} \underline{status} exists. Patient is suffering from \underline{hypothermia},
\end{quote}

\noindent
the set of gold terms would include all the underlined words. Among these words, \textit{infection}, \textit{altered}, \textit{mental}, \textit{status}, and \textit{hypothermia} are keyword terms, and \textit{no}, \textit{signs}, and \textit{of} are terms corresponding to the negation scope.

\subsubsection{Model:}
We split the dataset into subsets of 80-10-10\% as training-validation-test sets. We obtain a vocabulary of 47,015 tokens after lower-casing the documents without removing punctuation. We replace unknown words in validation and test sets with the $\langle$unk$\rangle$ token. We train LSTM classifiers to predict the document class from the hidden representation after the final timestep, which is obtained after processing the entire document as a sequence of tokens\footnote{We do not experiment with other types of classifiers because the focus of the work is to find and evaluate explanation rules for sequential models that use word embeddings as input, as opposed to comparing different classifiers.}. The classifiers use randomly initialized word embeddings and a single RNN layer without attention. The hidden state size and embedding dimension are set to either 50 or 100. We use the Adam optimizer~\citep{kingma2014adam} with learning rate 0.001 and a batch size of 64 (without hyperparameter optimization). Classification performance is shown in Table~\ref{tab:eval_quantitative}.

\subsection{Real clinical datasets}
\label{sec:clinical_dataset}
We additionally find explanation rules for sepsis classifiers on the MIMIC-III clinical corpus. We describe the definition of sepsis label in Appendix \ref{sec:sepsis_def}. 
\noindent 
We analyze two different setups after removing blank notes and the notes marked as \textit{error} in the MIMIC-III corpus:
\begin{enumerate}
    \item We use the last discharge note for every patient to classify whether the patient has sepsis. Class distribution among 58,028 instances is 90-10\% for non-septic and septic cases respectively, and the vocabulary size is 229,799. The task is easier in this setup because 70\% of septic cases mention sepsis directly, whereas only 13\% of non-septic cases mention sepsis. 
    \item We classify whether a patient has a sepsis diagnosis or not using the last note about a patient excluding the categories discharge notes, social work, rehab services and nutrition. We obtain 52,691 patients in this manner, out of which only 9\% are septic. The vocabulary size is 87,753. In this setup, only 17\% of septic cases mention sepsis, as opposed to 6\% of non-septic cases mentioning sepsis.
\end{enumerate}

\subsubsection{Models:} We train 2-layer bidirectional LSTM classifiers with 100 dimensional randomly initialized word embeddings and 100 dimensional hidden layer. We train for 50 epochs with early stopping with patience 5. The remaining data processing and implementation details are the same as discussed for synthetic dataset. Macro F1 score of classification when using discharge notes is 0.68 (septic class F1 is 0.41), and without using discharge notes is 0.60 (septic class F1 is 0.27). Majority baseline is 0.5.

\subsection{Sentiment analysis}
Following~\citet{DBLP:conf/iclr/MurdochS17}, we explain LSTM classifiers initialized with 300 dimensional Glove~\citep{pennington2014glove} embeddings and 150 hidden nodes for binary sentiment classification on the Stanford sentiment analysis dataset (SST2). We obtain 84.13\% classification accuracy, and our vocabulary size is 13,983.

\section{Evaluation}

\subsection{Comparing pooling techniques}
\label{sec:res_pooling}

To compare different pooling techniques described in Section~\ref{sec:pooling}, we evaluate sets of most important words obtained by different techniques against gold sets of important terms for the documents.

\subsubsection{Qualitative analysis}
\label{sec:eval_qualitative}

\begin{figure*}
    \begin{subfigure}{\linewidth}
    \centering
    \includegraphics[width=\linewidth]{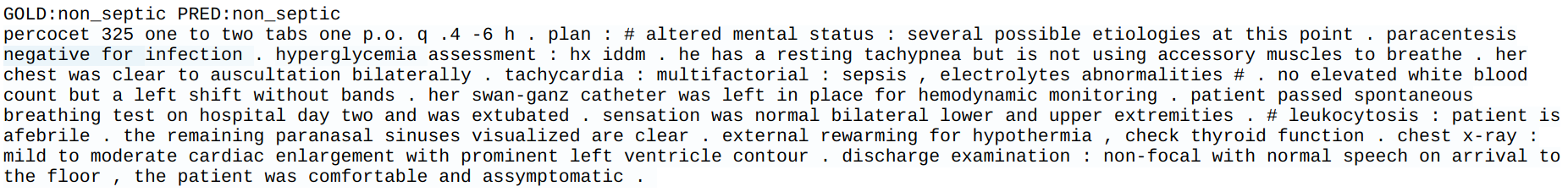}  
    \caption{L2 norm}
    \label{fig:l2}
    \end{subfigure}
    
    \begin{subfigure}{\linewidth}
    \centering
    \includegraphics[width=\linewidth]{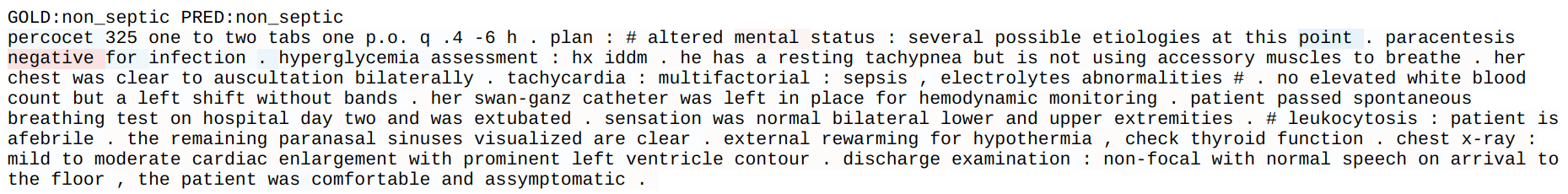}  
    \caption{Sum}
    \label{fig:sum}
    \end{subfigure}
    
    \begin{subfigure}{\linewidth}
    \centering
    \includegraphics[width=\linewidth]{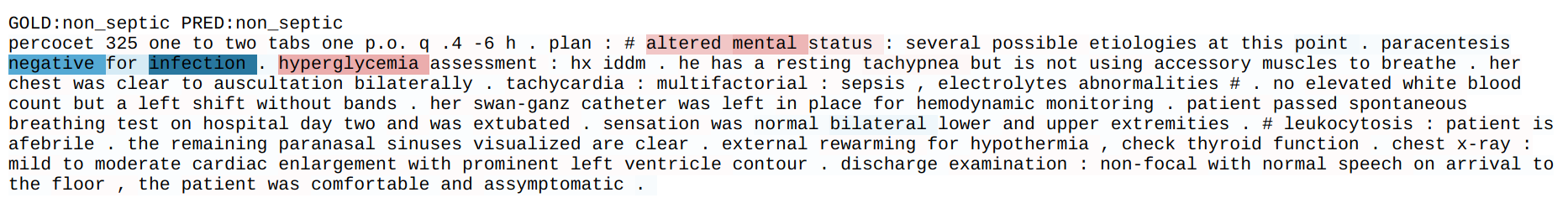}  
    \caption{Dot}
    \label{fig:dot}
    \end{subfigure}

    \caption{Heatmap visualization of word importance distribution for a single validation set instance in LSTM classifier with 50 hidden nodes and 100 dimensional word embeddings when L2, sum, and dot pooling techniques are used. Blue reflects positive importance and red indicates negative importance.}
    \label{fig:heatmaps}
\end{figure*}

In Figure~\ref{fig:heatmaps}, we compare word importance distribution for the pooling techniques for an instance in the validation set of the synthetic corpus. The L2 norm provides distributions over the positive values only and the importance scores are low because it squares the gradients. Sum pooling and dot product instead return a distribution over both positive and negative values, with dot product returning a more peaked distribution. However, as we can see, sum and dot product often provide opposite importance signs for the same words. This is caused due to presence of word embeddings while computing dot product, which can take both positive and negative values. In this instance, both true and predicted classes are \textit{non-septic}. Looking at Figure~\ref{fig:dot}, we find positive peaks over \textit{negative} and \textit{infection}, and negative peaks over \textit{altered mental status} and \textit{hyperglycemia}. This corresponds to the class labeling rule in the synthetic data, where \textit{non-septic} class is assigned when infection terms are negated. These directions of influence are counter-intuitive for sum pooling in Figure~\ref{fig:sum}. Due to its intuitive, peaked importance distributions, dot product seems to be better than other techniques. However, we move to quantitative evaluation for a global perspective because this qualitative analysis is biased towards a specific instance and model.

\subsubsection{Quantitative analysis}
\label{sec:eval_quantitative}
We find the top $k$ tokens for test documents in the synthetic dataset by ranking absolute word importance scores, where $k$ is the number of gold important terms used to label the document. We ignore the 20k documents that only consist of sentences that do not mention any keyword term, and hence have an empty gold set. We compute the accuracy of the set of most important words for every document compared to their corresponding gold set. Later, we take a mean across all the documents and report it in Table~\ref{tab:eval_quantitative}. We find that dot product consistently recovers more important tokens than other pooling techniques across all the classifiers, confirming the qualitative analysis earlier and the findings of~\citet{arras-etal-2019-evaluating}. Hence we use dot product for computing word importance before inducing explanation rules.

\begin{table}
\centering
\begin{tabular}{lr|rrr}
\hline 
\multicolumn{2}{c}{Classification} & \multicolumn{2}{r}{Pooling} \\
 Classifier & Acc. & L2 & sum & dot \\
\hline
LSTM100, E100 & 96.5 & 17.8 & 13.7 & \bf 26.0 \\
LSTM100, E50 & 95.5 &  23.7 & 21.5 & \bf 35.4 \\
LSTM50, E100 & 92.0 & 38.2 & 33.5 & \bf 50.2 \\
LSTM50, E50 & 92.4 &  26.5 & 25.1 & \bf 36.1 \\
\hline
\end{tabular}
\caption{\label{tab:eval_quantitative} Classification accuracy of different LSTM classifiers and the average accuracy for the top $k$ words in documents in the \textbf{synthetic dataset} obtained with L2, sum and dot product pooling techniques. LSTM$x$, E$y$ refers to LSTM with $x$ hidden nodes and $y$ dimensional word embeddings.}
\end{table}

We additionally see that the mean accuracy is nearly twice for the classifier with 50 hidden nodes and 100 dimensional word embeddings as compared to the the larger classifier that uses 100 hidden units instead, although the latter classifier is nearly 5\% more accurate. This suggests that the larger network obtains higher performance by focusing on tokens that are not incorporated within the gold keywords. The reason behind different tokens being considered important could be that our gold set of important terms is noisy:

\begin{itemize}
    \item Some tokens such as punctuation symbols are missing from the gold set, although they are important for identifying the scope of negation, as seen in Figure~\ref{fig:rules_synthetic}.
    \item Some terms in the gold set are not required for correct classification. For example:
    \begin{enumerate*} 
        \item Too many words are included as negation triggers. For example, in the sentence \textit{no signs of infection were found.}, `no', `signs', and `of' are all added to the gold set as negation markers although the subset \{`no',  `infection'\} may be sufficient. 
        \item Similarly, the keyword \textit{altered mental status} could already be recognized from a subset of these terms.
    \end{enumerate*}
    
\end{itemize}

\subsection{Baseline explanation rules}
Several existing approaches for global rule-based interpretability~\citep{lakkaraju2017interpretable,DBLP:journals/corr/PuriGAVK17} have one common aspect---they directly use the original input to find explanation rules for complex classifiers without making use of the parameters of the complex models. However, these approaches don't scale to NLP tasks due to combinatorial computational complexity in finding explanation rules. For comparison, as baseline rules, we induce explanations directly from the input data without using gradients of neural models. To this end, we create a bag-of-skipgrams by binarizing the most frequent skipgrams to represent whether they are present in a document. We then train rule induction classifiers on this binarized skipgram data to explain neural outputs.

We also compare to Anchors~\citep{anchors:aaai18} for SST2 explanations by implementing their submodular pick algorithm for obtaining global explanations. Anchors does not scale to longer documents used for sepsis classification. 

\subsection{Evaluation metrics}
We record fidelity scores of the explanation rules on the test set, and the complexity of these explanations. Fidelity scores refer to how faithful the explanations are to the test output predictions of the explained neural network. Like~\citet{W18-5411}, we use macro F1-measure of explanations compared to original predictions to quantify it. We define explanation complexity as the number of rules in an explanation.

\subsection{Explaining synthetic data classifiers}

\begin{table*}
\centering
\begin{tabular}{llrrrr}
 Explanation & Eval type & LSTM100,E100 & LSTM100,E50 & LSTM50,E100 & LSTM50,E50 \\
 \hline
\multirow{2}{*}{Baseline(sg)} & Fidelity & 75.65 & 77.67 & 83.19 & 84.30 \\
& Complexity & 63 & 60 & 26 & 46 \\
\multirow{2}{*}{\textsc{unravel}(sg)} & Fidelity & 98.90 & 99.46 & 99.97 & 98.24 \\
& Complexity & 32 & 13 & 2 & 49\\
\multirow{2}{*}{\textsc{unravel}(1gram)} & Fidelity & 98.83 & 99.51 & 99.97 & 97.22 \\
& Complexity & 23 & 18 & 2 & 51 \\
\hline
\end{tabular}
\caption{\label{tab:res_synthetic} Test set fidelity scores of explanations (in \%macro-F1), and number of explanation rules as the measure of explanation complexity for different LSTM classifiers on the \textbf{synthetic dataset} using our approach compared to the baseline approach. LSTM$x$,E$y$ refers to LSTM with $x$ hidden nodes and $y$ dimensional word embeddings.}
\end{table*}

We obtain explanations of all the LSTM classifiers for the synthetic dataset. We record fidelity scores of explanations and the corresponding complexity in Table~\ref{tab:res_synthetic}. We find that when we use the proposed pipeline \textsc{unravel} for learning gradient-informed rules, we obtain explanations with high fidelity scores also on the test data. On the other hand, with the baseline approach, we obtain nearly 15\% lower fidelity scores. In addition, explanations are more complex with the baseline approach. This confirms that making use of model parameters by means of gradients acts as an additional useful cue for the rule-based explainability module, thus resulting in more faithful explanations.

\begin{figure*}
\centering
\begin{subfigure}{\textwidth}
(a) \textbf{if} \textit{\underline{\underline{hyperglycemia}}} = $++$ \textsc{and} \textit{to exclude} = $0$ \textsc{and} \textit{evidence \underline{infection} .} = $0$ \textsc{and} \textit{\underline{infection}} = $++$ \textsc{and} \textit{no \underline{infection} .}= $0$ \textsc{and} \textit{no \underline{infection}} = $0$ \textsc{and} \textit{negative \underline{infection}} = $0$ \textsc{and} \textit{or of \underline{infection}} = $0$ \textsc{and} \textit{fungal \underline{infection} other} = $0$ \textsc{and} \textit{of \underline{infection} in the} = $0$ \textsc{and} \textit{\underline{\underline{altered}}} = $++$ 
$\implies$ septic (17466/17466)
\end{subfigure}
\begin{subfigure}{\textwidth}
\vspace{1em}%
(b) \textbf{if} \textit{\underline{\underline{tachypnea}}} = $0$ \textsc{and} \textit{\underline{meningitis}} = $0$ \textsc{and} \textit{urinary tract} = $0$ \textsc{and} \textit{\underline{endocarditis}} = $0$ \textsc{and} \textit{\underline{\underline{hyperglycemia}}} = $0$ 
$\implies$ non-septic (16015/16015)
\end{subfigure}
\begin{subfigure}{\textwidth}
\vspace{1em}%
(c) \textbf{if} \textit{no} = $++$ \textsc{and} \textit{urinary} = $0$ \textsc{and} \textit{bacterial} = $0$ \textsc{and} \textit{\underline{\underline{mental}}} = $-$ $\implies$ non-septic (1277/1345)
\end{subfigure}


\caption{Example explanation rules for the best \textsc{lstm} classifier on the \textbf{synthetic dataset}. Infection keywords from set $I$ are marked with a single underline, and the corresponding inflammatory response keywords from set $Infl$ are marked with double underline. $++$ refers to high positive importance of a term, $0$ represents absence of a term, and $-$ means that the term gets a low negative importance, i.e., presence of the term reduces the output probability. The numbers $(a/b)$ mean that $b$ training instances are explained by the rule, of which $a$ are correct. The first two rules are obtained with skipgrams, and the third one is obtained on using only unigrams for explanations.}
\label{fig:rules_synthetic}
\end{figure*}

\begin{table}
\centering
\begin{tabular}{llrr}
 Dataset & Explanation & Fidelity & N\_rules \\
 
 \hline

\multirow{2}{*}{+discharge} & Baseline(sg) & 61.7 & 825 \\
& \textsc{unravel}(sg) & 97.9 & 16 \\

\hline

-discharge & \textsc{unravel}(sg) & 77.3 & 196 \\
\hline
\multirow{2}{*}{SST2} & Anchors & 70.3 & 10 \\
& \textsc{unravel}(sg) & 80.2 & 87 \\
\hline
\end{tabular}
\caption{\label{tab:res_mimic} Explanation fidelity (\% macro F1) and complexity for \textbf{sepsis classification}: 1) With discharge notes 2) Without discharge notes, and on the \textbf{SST2} dataset. The baseline method did not converge (in several weeks) for sepsis classification without discharge note and for SST2 classification. Anchors did not scale (in memory usage) to document-level sepsis datasets.}
\end{table}

We present some examples of explanation rules for the most accurate LSTM classifier for the synthetic dataset in Figure~\ref{fig:rules_synthetic}. Here, we indicate infection keywords that were used to populate the dataset with a single underline, and the inflammatory response keywords with a double underline. The first rule in the figure indicates that if two inflammatory response criteria are highly important for the network, the term infection is highly important, and phrases negating the presence of infection are absent, then the class is recognized as septic. This is similar to the rule we have used to label the synthetic dataset, which requires at least one infection term and at least two inflammatory response criteria to not be negated in the document for being assigned a septic class. In the next rule---applied after all the cases from the previous rule have been excluded from the dataset---if several keyword terms are absent, the document is classified as non-septic. It is useful to remember that \textit{urinary tract} is usually followed by the word \textit{infection} in the dataset, and several instances mentioning \textit{infection} have already been explained by the previous rule and hence have been ignored by this rule. This explanation rule is also in accordance to the synthetic dataset, where 20k documents do not contain any keyword term and are labeled as non-septic.

The third rule is an example rule for the same model when explanations are based on unigrams only as opposed to skipgrams. In this case, we lose the context of the negation marker \textit{no}. When using skipgrams, this context of negation is available, which makes the negation scope clearer. Further, terms like \textit{evidence}, \textit{fungal} and \textit{urinary tract} captured by skipgrams provide additional context for understanding the rules. This illustrates that even though the fidelity scores of explanations are similar, skipgram based explanations are more interpretable than only unigram-based explanations. Hence, we use skipgrams for further analysis.

\begin{figure}

\textbf{if} \textit{sepsis major surgical} = ++ $\implies$ septic (209/209) \\ 
\textbf{if} \textit{complaint : sepsis} = 0 \textsc{and}
 \textit{chief hypotension major} = ++ $\implies$ septic (169/169)

\caption{First two explanation rules for the \textbf{clinical dataset} that uses discharge notes to classify sepsis. $++$ refers to high positive importance, and $0$ refers to an absent term in the document. $(a/b)$ in parentheses show that $a$ of $b$ examples explained by this rule are correct.}
\label{fig:rules_mimic_discharge}
\end{figure}

\subsection{Explaining clinical models}

We rerun our explainability pipeline on both clinical models for sepsis classification---with and without using discharge notes (Section~\ref{sec:clinical_dataset}). 
For the first classifier with discharge notes, we again obtain very high fidelity scores of explanations (Table~\ref{tab:res_mimic}). The baseline explanations have significantly lower fidelity scores while also being extremely complex. On inspecting the corresponding explanation rules given in Figure~\ref{fig:rules_mimic_discharge}, we find that they refer to the direct mentions of sepsis in the discharge notes. In the first rule, if \textit{sepsis major surgical} is mentioned, the class is directly septic. In the second condition, it first rules out the mention of a complaint of sepsis and then checks for additional conditions. This confirms that not only does the classifier pick up on these direct mentions, but the explanations also recover this information. This illustrates the utility of \textsc{unravel} in understanding our models, which is the first step towards improving them. For example, if our model is learning direct mentions of sepsis as a discriminating feature, we could remove these direct mentions from the dataset before training new models to ensure that they generalize.    

Next, for the more difficult case where we use only the final non-discharge note about patients to classify whether they have sepsis, the fidelity score is 77.33\%. Although this score is good as an absolute number, it is much lower than other two cases. Explanations for this model are also much more complex. This highlights that more complex classifiers and explanations have lower explanation fidelity. While manually inspecting these explanations, we find that absence of terms such as \textit{diagnosis : sepsis}, \textit{indication endocarditis . valve}, \textit{indication bacteremia}, \textit{admitting diagnosis fever} and \textit{pyelonephritis} are used to rule out sepsis. These are similar to the explanations of the other two datasets, albeit enriched with information about additional infections and body conditions. This confirms that the synthetic dataset closely models a real clinical use case, and suggests that these explanations rules could result into useful hypothesis generation. 

\subsection{Explaining sentiment classifier}

Results of the \textsc{SST2} explanations are given in Table~\ref{tab:res_mimic}. Our pipeline provides $\scriptstyle\mathtt{\sim}$10\% more accurate explanations compared to Anchors. Moreover, on inspecting the rules (presented in Appendix \ref{sec:sst2_rules}), we find that Anchors rules consist only of single words, as opposed to \textsc{unravel}, which finds conjunctions of different phrases. Furthermore, explanation rules with \textsc{unravel} obtain 71\% classification accuracy on the original task. This performance drop compared to LSTM is $\scriptstyle\mathtt{\sim}$7\% lower than gradient decomposition-based performance drop reported by~\citet{DBLP:conf/iclr/MurdochS17}, although the numbers aren't strictly comparable because we explain different classifiers\footnote{Their implementation is not openly available for direct comparison.}.

\section{Conclusions and Future Work}
We have successfully developed a pipeline to obtain transferable, accurate gradient-informed explanation rules from RNNs. We have constructed a synthetic dataset to qualitatively and quantitatively evaluate the results, and we obtain informative explanations with high fidelity scores. We obtain similar results on clinical datasets and sentiment analysis. Our approach is transferable to all similar neural models. In future, it would be interesting to extend the capabilities of this approach to obtain more accurate, less complex and scalable explanations for classifiers with more complex patterns.

\section*{Acknowledgments}
This research was carried out within the Accumulate strategic basic research project, funded by the government agency Flanders Innovation \& Entrepreneurship (VLAIO) [grant number 150056].

\bibliography{acl2020}
\bibliographystyle{acl_natbib}

\appendix

\section{Sepsis Definition}
\label{sec:sepsis_def}
We define sepsis label as all the cases where patients are assigned one of the following diagnostic codes: 
\begin{itemize}
    \item 995.91 (Sepsis): Two or more systemic inflammatory response criteria plus a known or suspected infection. ~2\% of the cases.
    \item 995.92 (Severe Sepsis):
Sepsis with acute organ dysfunction. ~3\% of the cases.
    \item 785.52 (Septic Shock): Form of severe sepsis where the organ dysfunction involves the cardiovascular system. ~4\% of the cases.
\end{itemize}

\section{Explanation rules for SST2 classifier}
\label{sec:sst2_rules}

\begin{figure*}[htbp!]
    \centering
    \begin{subfigure}{\textwidth}
    \textbf{if} \textit{?} = 0 \textsc{and} \textit{bad .} = 0 \textsc{and} \textit{too} = $++$ \textsc{and} \textit{one} = 0 $\implies$ negative (159/159)
    \end{subfigure}
    
    \begin{subfigure}{\textwidth}
    \vspace{1em}%
    \textbf{if} ? = $++$ $\implies$ negative (81/82)
    \end{subfigure}
    
    \begin{subfigure}{\textwidth}
    \vspace{1em}%
    \textbf{if} bad . = 0 \textsc{and} worst = 0 \textsc{and} fails = 0 \textsc{and} feels = $++$ $\implies$ negative (54/54)
    \end{subfigure}
    
    \begin{subfigure}{\textwidth}
    \vspace{1em}%
    \textbf{if} bad . = 0 \textsc{and} worst = 0 \textsc{and} fails = 0 \textsc{and} is bad = 0 \textsc{and} flat = 0 \textsc{and} mess = 0 \textsc{and} stupid = 0 \textsc{and} \textit{suffers} = 0 \textsc{and} \textit{pointless} = 0 \textsc{and} \textit{dull} = $++$ $\implies$ negative (38/38)
    \end{subfigure}
    
    \begin{subfigure}{\textwidth}
    \vspace{1em}%
    \textbf{if} \textit{bad .} = $++$ $\implies$ negative (36/36)
    \end{subfigure}
    \caption{Example explanation rules for the \textbf{SST2 dataset}. $++$ refers to high positive importance, and $0$ refers to an absent term in the document. $(a/b)$ in parentheses show that $a$ of $b$ examples explained by this rule are correct.}
    \label{fig:rules_sst2}
\end{figure*}

In Figure~\ref{fig:rules_sst2}, we present the first explanation rules on explaining the LSTM classifier trained on the SST2 dataset with the \textsc{unravel} pipeline. In Figure~\ref{fig:sst2_anchors}, we present the explanations on the same dataset produced with the Anchors submodular pick algorithm.

\begin{figure*}
    \centering
    
    \begin{subfigure}{\textwidth}
    \vspace{1em}%
    \textbf{if} \textit{the} is present $\implies$ negative
    \end{subfigure}
    
    \begin{subfigure}{\textwidth}
    \vspace{1em}%
    \textbf{if} \textit{a} is present $\implies$ positive
    \end{subfigure}
    
    \begin{subfigure}{\textwidth}
    \vspace{1em}%
    \textbf{if} \textit{civility} is present $\implies$ positive
    \end{subfigure}
    
    \begin{subfigure}{\textwidth}
    \vspace{1em}%
    \textbf{if} \textit{of} is present $\implies$ positive
    \end{subfigure}
    
    \begin{subfigure}{\textwidth}
    \vspace{1em}%
    \textbf{if} \textit{this} is present $\implies$ negative
    \end{subfigure}
    
    \begin{subfigure}{\textwidth}
    \vspace{1em}%
    \textbf{if} \textit{just} is present $\implies$ negative
    \end{subfigure}
    
    \begin{subfigure}{\textwidth}
    \vspace{1em}%
    \textbf{if} \textit{good} is present $\implies$ positive
    \end{subfigure}
    
    \begin{subfigure}{\textwidth}
    \vspace{1em}%
    \textbf{if} \textit{with} is present $\implies$ positive
    \end{subfigure}
    
    \begin{subfigure}{\textwidth}
    \vspace{1em}%
    \textbf{if} \textit{no} is present $\implies$ negative
    \end{subfigure}
    
    \begin{subfigure}{\textwidth}
    \vspace{1em}%
    \textbf{if} \textit{little} is present $\implies$ positive
    \end{subfigure}
    
    \caption{Explanation rules for the LSTM classifier on the SST2 dataset with the Anchors submodular pick algorithm. The rules check the presence of words in the input to map to an output class.}
    \label{fig:sst2_anchors}
\end{figure*}

\end{document}